\documentclass[acmsmall, sigconf,screen,nonacm]{acmart}

\usepackage{acronym}
\usepackage{microtype}
\usepackage{pifont}
\usepackage{tabularx}
\usepackage{multirow}
\usepackage{amsmath}

\DeclareMathOperator*{\argmax}{argmax}
\DeclareMathOperator*{\argmin}{argmin}

\usepackage{acronym}

\acrodef{IoT}{Internet of Things}
\acrodef{NN}{Neural Network}
\acrodef{EE}{Early Exit}
\acrodef{EENN}{Early Exit Neural Network}
\acrodef{ECG}{Electrocardiography }
\acrodef{DL}{Deep Learning}
\acrodef{MCU}{Microcontroller}
\acrodef{DoS}{Denial-of-Service}
\acrodef{RNN}{Recurrent Neural Network}
\acrodef{MAC}{Multiply-Accumulate}
\acrodef{BMBF}{German Federal Ministry of Education and Research}
\acrodef{p.p.}{percentage point}
\acrodef{MI}{myocardial infarction}

\acrodef{DD}{Difference Detection}
\acrodef{TP}{Temporal Patience}
\acrodef{RDM}{Range-Doppler Map}

\begin{document}

\title{Temporal Decisions: Leveraging Temporal Correlation for Efficient Decisions in Early Exit Neural Networks
}

\author{Max Sponner}
\email{max.sponner@infineon.com}
\orcid{0000-0002-4830-9440}
\affiliation{%
  \institution{Infineon Technologies Dresden GmbH \& Co. KG}
  \streetaddress{Königsbrücker Str. 180}
  \city{Dresden}
  \country{Germany}
  \postcode{01099}
}

\author{Lorenzo Servadei}
\email{lorenzo.servadei@tum.com}
\orcid{0000-0003-4322-834X}
\affiliation{%
  \institution{Chair for Design Automation - Technical University of Munich}
  \streetaddress{Arcisstrasse 21}
  \city{Munich}
  \country{Germany}
  \postcode{80333}
}

\author{Bernd Waschneck}
\email{bernd.waschneck@infineon.com}
\orcid{0000-0003-0294-8594}
\affiliation{%
  \institution{Infineon Technologies AG}
  \streetaddress{Am Campeon 1}
  \city{Neubiberg}
  \country{Germany}
  \postcode{85579}
}

\author{Robert Wille}
\email{robert.wille@tum.com}
\orcid{0000-0002-4993-7860}
\affiliation{%
  \institution{Chair for Design Automation - Technical University of Munich}
  \streetaddress{Arcisstrasse 21}
  \city{Munich}
  \country{Germany}
  \postcode{80333}
}

\author{Akash Kumar}
\email{akash.kumar@tu-dresden.de}
\orcid{0000-0002-4830-9440}
\affiliation{%
  \institution{Chair of Processor Design, CfAED - Technical University of Dresden}
  \streetaddress{Helmholtzstr. 18}
  \city{Dresden}
  \country{Germany}
  \postcode{01069}
}

\begin{abstract}

    Deep Learning is becoming increasingly relevant in Embedded and Internet-of-things applications.
    However, deploying models on embedded devices poses a challenge due to their resource limitations.
    This can impact the model's inference accuracy and latency.
    One potential solution are Early Exit Neural Networks, which adjust model depth dynamically through additional classifiers attached between their hidden layers.
    However, the real-time termination decision mechanism is critical for the system's efficiency, latency, and sustained accuracy.

    This paper introduces Difference Detection and Temporal Patience as decision mechanisms for Early Exit Neural Networks.
    They leverage the temporal correlation present in sensor data streams to efficiently terminate the inference.
    We evaluate their effectiveness in health monitoring, image classification, and wake-word detection tasks.
    Our novel contributions were able to reduce the computational footprint compared to established decision mechanisms significantly while maintaining higher accuracy scores.
    We achieved a reduction of mean operations per inference by up to 80\,\% while maintaining accuracy levels within 5\,\% of the original model.
    
    These findings highlight the importance of considering temporal correlation in sensor data to improve the termination decision.
\end{abstract}

%

\maketitle

\section{Introduction}

Deploying Deep Learning inference workloads on \ac{IoT} and embedded devices presents significant challenges.
Deep Learning models have high computational demands, which leads to high power consumption and shortened battery life.
Additionally, constrained computational resources of such devices lead to prolonged processing latencies.
These limitations affect user experience.
While static optimizations, such as pruning and quantization, offer the potential to reduce the model's resource footprint, they also reduce the model's predictive capabilities permanently.

A possible solution are \acp{EENN}.
They are able to perform the trade-off between efficiency and prediction quality at runtime.
This is achieved by adding early classifiers between the model's hidden layers.
All of these classifiers are trained to perform the same task.
During the inference, the process can be terminated early if the result produced by an \ac{EE} is sufficient for the current situation.
However, the main challenge in deploying \acp{EENN} is the decision mechanism that governs the termination at runtime.
Their choice has a major influence on sustained accuracy and achieved efficiency gains.
Finding the ideal termination mechanism is an ongoing research area for \acp{EENN}.


State-of-the-art solutions in this area include rule-based strategies.
These rely on derived metrics such as the available compute budget~\cite{amthorImpatientDNNsDeep2016, huLearningAnytimePredictions2018} or the confidence~\cite{pandaConditionalDeepLearning2016, parkBigLittleDeep2015b} of \acp{EE}.
More sophisticated approaches involve employing a dedicated agent~\cite{bolukbasiAdaptiveNeuralNetworks2017a, odenaChangingModelBehavior2017} to determine when to terminate based on the current input sample.

This paper evaluates novel methods for the adaptive termination of \acp{EENN} that operate by monitoring output changes while processing temporally correlated samples.
They capitalize on the observation that similar inputs will create similar output patterns with minor deviations.
As long as the variation in output remains below a predefined threshold, the methods assume the input data to be fundamentally similar.
By terminating inference based on these monitored variations, the approaches aspire to optimize the performance of \acp{EENN} within resource-constrained environments.

The approaches demonstrate multiple advantages over the state-of-the-art when addressing temporally correlated data:
\begin{itemize}
    \item \textbf{Leveraging Temporal Correlation:}
    They leverage temporal correlation within sensor data to guide inference termination.
    The approaches efficiently determine when to halt predictions by considering the similarity between subsequent samples.
    This optimizes computational efficiency by utilizing \acp{EE} that might not yield high-quality predictions while maintaining overall prediction quality.
    \item \textbf{Universal Similarity Metric:}
    The approaches use the \ac{EE} outputs as embeddings.
    This removes the need for domain-specific similarity metrics, establishing a universal metric across diverse data modalities and tasks.
    By quantifying similarity through \ac{EE} outputs, we adapt it to unique data characteristics and leverage features learned by the underlying \acp{NN} for specific tasks.
    
    \item \textbf{Enhanced Efficiency via Scene Grouping:}
    The methods introduce a mechanism for detecting sequences of similar samples.
    The identification of such sequences enhances the decision process resilience against gradual changes in input data.
    Additionally, this information could be used for monitoring or similar tasks in future work.
\end{itemize}

The approaches highlight the potential of utilizing the temporal correlation in sensor data to enhance the quality of \acp{EENN} termination decisions.


\section{Related Work}

\begin{figure}
    \centering
    \includegraphics[width=0.95\linewidth]{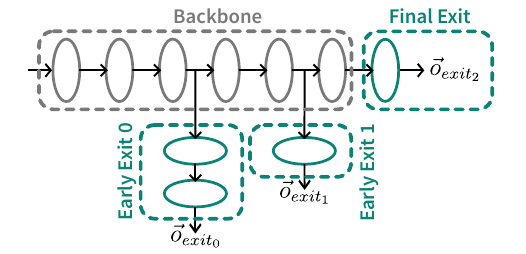}
    \caption{An Early Exit Neural Network with two Early Exit branches.
    Each branch contains an output that produces a classification vector of the same shape as the final classifier, as all outputs are intended to solve the same task.}
    \label{fig:eenn}
\end{figure}

\Acp{EENN} were introduced with BranchyNet by Teerapittayanon in 2016~\cite{teerapittayanonBranchyNetFastInference2016a}.
These \acp{NN} incorporate \aclp{EE} (\acp{EE}) -- additional classifier branches located between the network's hidden layers (see Fig.~\ref{fig:eenn}).
\acp{EE} perform the same task as the original classifier but only utilize features extracted up to their attachment point in the network architecture.

This allows for inference termination when an \ac{EE}'s output suffices.
Avoiding deeper layer execution enhances latency and power efficiency and can improve model accuracy~\cite{bolukbasiAdaptiveNeuralNetworks2017a, odenaChangingModelBehavior2017}.
The challenge is maintaining predictive capabilities while achieving efficiency gains through the runtime decision mechanism.

Several decision mechanisms have been explored.
Rule-based approaches are among the most resource-efficient solutions.
They rely on a dynamic compute budget or the confidence levels from encountered classifiers to determine termination. 
Budget-based decisions can be made a-priori~\cite{amthorImpatientDNNsDeep2016} or just-in-time~\cite{huLearningAnytimePredictions2018}, depending on when the budget information is available.
The a-priori method is especially efficient as it executes only the last classifier that fits within the budget.
However, in scenarios where the budget is unknown beforehand, all classifiers must be executed until the resources run out.
A drawback of this approach is that it only relies on resource information -- overlooking the complexity of the processed data.
This can lead to overspending on simpler data and reduce accuracy on more complex samples.

An alternative that integrates the input complexity is the con\-fi\-dence-based decision mechanism~\cite{pandaConditionalDeepLearning2016, parkBigLittleDeep2015b}.
This technique derives metrics (confidence, score margin, or entropy) from an \ac{EE}'s output, which are used to estimate result correctness.
However, this approach mandates always executing all encountered \ac{EE} branches before termination and lacks a-priori decision planning.
Moreover, the derived metrics may not reliably correlate with the correctness of the output~\cite{nguyenDeepNeuralNetworks2015}, leading to sub-optimal results.

For deeper networks with many \acp{EE}, the patience approach~\cite{zhouBERTLosesPatience2020} emerges.
It compares the output of the most recently encountered \ac{EE} classifiers and halts the inference if they agree on the output.
This method has mostly been used for natural language processing tasks~\cite{zhouBERTLosesPatience2020}.

Template-matching~\cite{rashidTemplateMatchingBased2022} combines domain knowledge and con\-fi\-dence-based \acp{EENN}.
This is achieved by incorporating the confidence of the \ac{EE} classifier alongside the similarity to a reference sample.
This adds domain-specific knowledge to the decision process.
However, the similarity metric must be able to be computed within the limited time window for the at-runtime decision and is often only applicable to certain data modalities.

Other solutions prioritize prediction quality over improved efficiency.
They incorporate an additional \ac{NN} as an agent for terminating supervised \acp{EENN}~\cite{bolukbasiAdaptiveNeuralNetworks2017a, odenaChangingModelBehavior2017}.
While this approach improves the accuracy by directly considering the input data for its decision, its increased resource footprint makes it impractical for embedded scenarios.

While current solutions have improved the performance of \acp{NN}, they often fail to leverage the temporal correlations found in most sensor data.
To address this issue, the novel decision mechanisms aim to utilize the correlations present in streaming sensor data.
By doing so, the approach can lead to further improvements in processing this type of data with \acp{EENN}.

A previous feasibility study (preprint)~\cite{sponner2023temporal} evaluated the effectiveness of this method for radar data, demonstrating an improvement in efficiency while maintaining accuracy.
Due to the high sampling rate, radar data streams are highly correlated, as environmental changes happen much slower.
Building on this, our current research aims to generalize and expand upon these findings to different data modalities with temporal correlations and extend it with an alternative scene labeling mechanism.  



\section{Methodology}


The proposed decision mechanisms capitalize on the temporal correlation inherent in sensor data and the propagation of changes throughout the network architecture to facilitate efficient termination decisions.

In classification scenarios involving multiple classes, the \ac{EE} classifier generates a vector representing the detected classes.
Each vector element denotes the estimated probability of the input sample belonging to the class indexed according to the \ac{NN}.
These results establish a vector space suitable for distance calculation.
To evaluate the change between subsequent samples, we compute the Euclidean distance ($\operatorname{d}_{\text{euclidean}}$) between the results ($\vec{o}_{t_i,\text{exit}_n}$) produced by an \ac{EE}  classifier of the \ac{NN} (see Eq.~\ref{eq:change_metric}).
Alternatively, other distance metrics can be incorporated.
For binary classification or regression tasks, a straightforward scalar difference suffices.
Although the method can be applied to segmentation tasks, the larger output tensor's computational demands might offset the gains achieved by the early termination of the inference.

\begin{equation}
\small
\operatorname{change}(t_1, t_{\text{initial}}) = \operatorname{d}_{\text{euclidean}}(\vec{o}_{t_1,\text{exit}_0},\vec{o}_{t_{\text{initial}},\text{exit}_0}) \quad , \quad \vec{o} \in \mathbb{R}^C
\label{eq:change_metric}
\end{equation}

This approach leverages \ac{EE} outputs as semantic embeddings based on extracted features up to the utilized classifier.
These embeddings are not expressive enough to reconstruct the original inputs but can be used to compare them within the limited set of recently encountered samples.
The \acp{EE} are trained to perform the network's task on the input modality, which automatically includes the necessary information for the current modality and task.
This eliminates the need for handcrafted similarity metrics.
Utilizing \ac{EE} outputs for similarity calculation enables the sharing of computations between inference and similarity detection, enhancing efficiency.
Additionally, as the network output is typically smaller than the input, this method can have a smaller memory footprint compared to approaches incorporating input similarity into the decision-making process.
Once the similarity between the current and reference samples in the \ac{EE} classifier output space is calculated, it is compared to a predefined threshold.
This threshold is a hyperparameter introduced by the decision mechanism that needs to be configured by a developer, similar to the thresholds used in confidence-based solutions.

\begin{figure}
    \centering
    \includegraphics[width=1.0\linewidth]{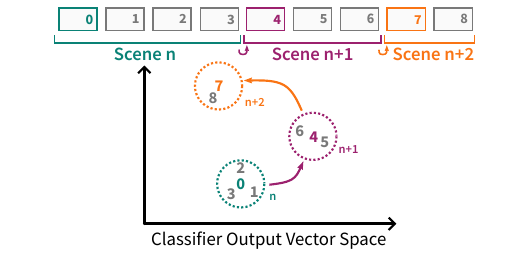}
    \caption{Scenes are detected by calculating the distance of an \ac{EE} classifier's output to a reference predecessor.
    Distances above the preset threshold indicate the start of a new scene, otherwise the current scene continues.
    }
    \label{fig:temporal_patience}
\end{figure}

Rather than using the direct predecessor as a reference -- which could lead to gradual change accumulation and inaccurate labeling -- the solution aims to categorize consecutive inputs into ``scenes".
A scene consists of a sequence of similar samples whose change remains below the acceptable threshold. 
The first sample of a scene is identified by being too different from the previous reference and then designated as the new reference for ensuing inputs (see Fig.~\ref{fig:temporal_patience}).
The reference samples receive labels through the majority vote ($\operatorname{vote}({o_{t,\text{exit}_0}, o_{t,\text{exit}_1}, ..., o_{t,\text{exit}_n}})$) of all \ac{NN} classifiers  (see Eq.~\ref{eq:majority_vote}).
The majority vote is intended to prevent overthinking -- a phenomenon where the deeper classifiers overwrite the correct output of an \ac{EE}~\cite{kayaShallowdeepNetworksUnderstanding2019}.
As this behavior is not present in all \acp{EENN}, we also evaluate the alternative of only using the last classifier to label a new scene (see Section~\ref{sec:new_scene}).
\begin{equation}
\small
\operatorname{vote}({o_{t,\text{exit}_0}, o_{t,\text{exit}_1}, ..., o_{t,\text{exit}_n}}) = \argmax_{c \in C} \left(\sum_{i=1}^{n} [o_{t,\text{exit}_i} = c]\right)
\label{eq:majority_vote}
\end{equation}

Subsequent samples within a scene solely obtain labels from the classifier utilized for assessing their similarity to the reference sample.
Although both the Difference Detection and Temporal Patience decision algorithms adopt this approach, they differ in how they select the employed \ac{EE} classifier, label subsequent scene samples, and recognize the onset of a new scene.

\subsection{Difference Detection}

In the context of Difference Detection, the initial \ac{EE} classifier within the \ac{NN} is always used to quantify similarity.
This minimizes the cost of processing the subsequent samples of a scene.
Detecting a new scene relies exclusively on the calculated distance value at this specific classifier (see Eq.~\ref{eq:change_metric}).
If the difference between a sample and the previous scene surpasses a preset threshold, the mechanism invokes the majority vote of all classifier outputs for that sample.
When the difference falls below the threshold, the label from the preceding majority vote is reused and the inference terminates early.
This implementation aligns with input filtering strategies, albeit employing the early result's distance instead of a manually crafted similarity metric.
Refer to Eq.~\ref{eq:difference_detection} for a comprehensive representation of the complete Difference Detection process.
\begin{equation}
\small
\operatorname{output}(t) = 
  \begin{cases}
    \operatorname{vote}_{t_{\text{initial}}} & \text{,if } \operatorname{change}(t, t_{\text{initial}}) < \mathrm{threshold} \\
    \operatorname{vote}_{t} & \text{,if } \operatorname{change}(t, t_{\text{initial}}) \ge \mathrm{threshold}
  \end{cases}
\label{eq:difference_detection}
\end{equation}

\subsection{Temporal Patience}

Using only the first \ac{EE} can limit the accuracy, as shallower hidden layers might fail to extract features that are relevant for the correct classification of the current sample. 
This constraint prevents the classifier from capturing changes in these crucial features that are only extracted by deeper layers.
To enhance accuracy, the selection of the Difference Detection classifier becomes variable.
Instead of exclusively using the shallowest \ac{EE}, the shallowest classifier that agrees with the majority vote of the scene's initial sample (see Eq.~\ref{eq:tp:selection}) determines the similarity for subsequent inputs within the scene (see Eq.~\ref{eq:tp:change_metric}).
Important to note is that distance calculations will always happen between the outputs of the same \ac{EE}.

\begin{equation}
\small
\operatorname{select}(t) := \underset{i}{\argmin} ( o_{t,\text{exit}_i} = \operatorname{vote}(o_{t,\text{exit}_0}, o_{t,\text{exit}_1}, ..., o_{t,\text{exit}_n}))
\label{eq:tp:selection}
\end{equation}

This approach enhances accuracy by ensuring that the chosen classifier can effectively extract the necessary features to label the initial sample accurately.
While it may increase the average inference cost due to the execution of additional layers for deeper \acp{EE}, it is intended to maintain the accuracy more effectively.

\begin{equation}
\small
\operatorname{change}(t, t_{\text{initial}}) = \operatorname{d}(\vec{o}_{t, \text{exit}_i},\vec{o}_{t_{\text{initial}},\text{exit}_i}) ~ , ~ i = \operatorname{select}(t_{\text{initial}})
\label{eq:tp:change_metric}
\end{equation}


The scene change detection incorporates the output class label to further enhance accuracy.
Detecting a new scene is determined not only by the distance threshold but also by a change in class label, which is the index of the highest value in the output vector. 
This reduces dependency on the threshold hyperparameter and mitigates the impact of incorrect configurations on system efficiency and accuracy.
The additional overhead is minimal, as the used \ac{EE} is already utilized for distance calculation.
The Temporal Patience decision mechanism, which introduces variable \ac{EE} selection and class label consideration, can be expressed by Eq.~\ref{eq:temporal_patience} and is designed to enhance accuracy compared to the Difference Detection mechanism.


\begin{equation}
\small
\begin{aligned}
\operatorname{update}(t) &= \operatorname{change}(t, t_{\text{initial}}) < \mathrm{threshold} \quad \text{and} \\
&\quad ~ \argmax_{c \in C}(\vec{o}_{t,\text{exit}_i}) = \argmax_{c \in C}(\vec{o}_{t_\text{initial},\text{exit}_i}) \\
\operatorname{output}(t) &=
\begin{cases}
\vec{o}_{t,\text{exit}_i} & \text{, if } \operatorname{update}(t) \\
\operatorname{vote}(o_{t,\text{exit}_0}, o_{t,\text{exit}_1}, \dots, o_{t,\text{exit}_n}) & \text{, otherwise}
\end{cases}
\end{aligned}
\label{eq:temporal_patience}
\end{equation}

\section{Evaluation and Discussion}

We evaluated the decision mechanisms across edge and \ac{IoT} applications involving temporally correlated sensor data with a range of possible decision thresholds.
If \acp{EENN} have previously been used in these scenarios, we adopted their architectures.
Otherwise, common single-exit architectures were converted into \acp{EENN}.
For test sets without temporal correlation, we augmented them using diverse methods to establish correlations between samples.
Our evaluation utilized accuracy and mean \ac{MAC} operations per inference as primary metrics.
The benefit of using \ac{MAC} operations is the hardware-independence of the metric:
a reduction in mean \ac{MAC} operations will translate to shorter latencies and reduced energy consumption across all hardware architectures.

\subsection{Mycordial Infarction Detection on ECG Data}

Our first benchmark aimed to detect \acp{MI} from single-lead \ac{ECG} data using the PTB-XL dataset~\cite{wagner2020ptb}.
Using single-lead ECG data enables wearable devices like smartwatches to alert users of acute health problems like \acp{MI}.
\Acp{MI} are a major health concern that require fast care, underscoring the critical need for fast and accurate detection methods. 
Efficient processing is crucial for wearable devices, ensuring continuous monitoring while maintaining long battery life.

We used the \ac{EENN} architecture with a single \ac{EE} and the preprocessing steps from a previous study~\cite{rashidTemplateMatchingBased2022}.
Both classifiers were jointly trained, resulting in an accuracy of 80.5\,\% for the final classifier and the single-exit reference version.
The \ac{EE} on its own achieved an accuracy of 78.8\,\%.
The computational load of the reference model without \acp{EE} is 16,776 \ac{MAC} operations.
The entire \ac{EENN} inference requires 16,957 \ac{MAC} operations, with only 181 of these required to execute the additional \ac{EE} branch.
The template-matching mechanism from the previous study~\cite{rashidTemplateMatchingBased2022} requires an additional 1,057 operations to compute the input similarity.

\begin{figure}[ht]
\centering
\fontsize{10}{12}\selectfont
\includegraphics[width=1.0\linewidth]{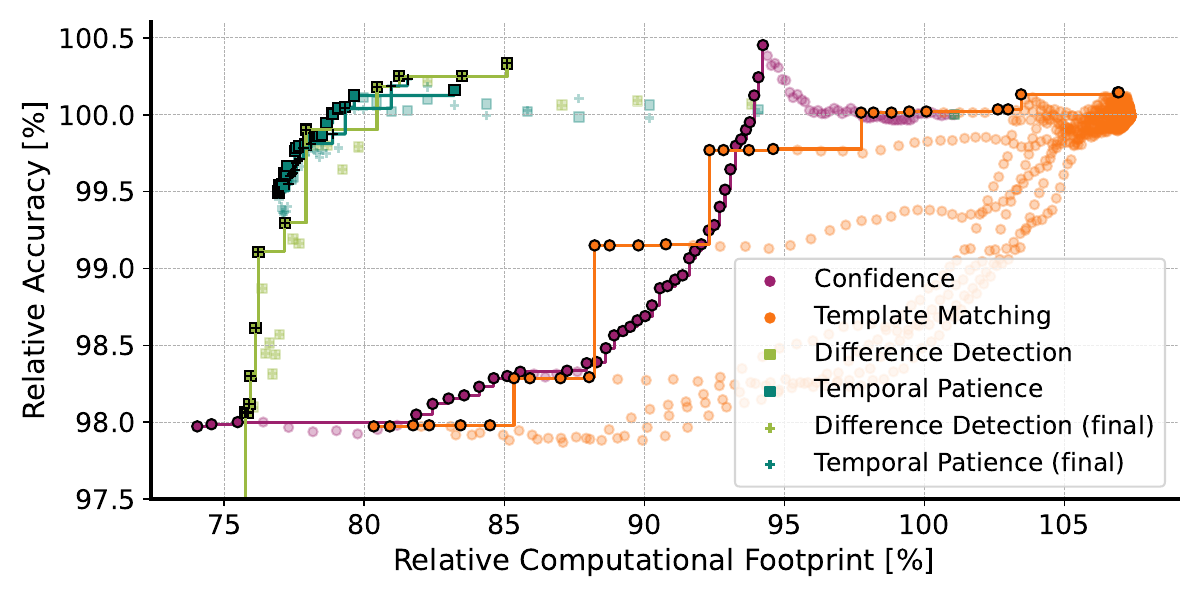}
\caption{
Mean operations per inference vs. accuracy on the PTB-XL test set for the different decision methods across different threshold configurations relative to the single exit version of the model.
}
\label{fig:ecg_scatter}
\end{figure}

We evaluated confidence-based and template-matching~\cite{rashidTemplateMatchingBased2022} alongside our approach using the same \ac{EENN}.
Our evaluation focused on accuracy and computational cost relative to the single-exit version (see Fig.~\ref{fig:ecg_scatter}).
The Difference Detection and Temporal Patience mechanisms exhibited similar accuracy levels with reduced computational footprints compared to existing solutions.
Optimal configurations of these mechanisms demonstrated 10-20\,\% fewer computations compared to other \ac{EENN} mechanisms.
Compared to the single-exit reference, a reduction of up to 20\,\% at equivalent accuracy levels was achieved.
This efficiency enhancement was achieved by the decreased usage of the final classifier and increased labeling by the \ac{EE}, capitalizing on the sample similarity within each \ac{ECG} recording.

The Difference Detection solution's accuracy was highly sensitive to the threshold hyperparameter, where an improper configuration could significantly decrease accuracy below that of just the \ac{EE} classifier.
In contrast, the Temporal Patience mechanism showed robustness against this effect, consistently achieving higher accuracy across various configurations.
This was achieved by the mechanism's change detection, which incorporated the class label of the \ac{EE} to reduce reliance on the threshold hyperparameter.
As a result, the Temporal Patience mechanism consistently outperformed the accuracy achievable by the \ac{EE} classifier on its own.

The initial experiment showed that temporal mechanisms in \acp{EENN} could provide efficiency advantages without compromising accuracy.
In domains prioritizing precision, such as medical applications, the Temporal Patience solution emerged as the preferred mechanism, maintaining accuracy while surpassing state-of-the-art alternatives in efficiency.



\noindent

\subsection{Image Classification}

In the second experiment, we evaluated the approach on an image classification task by augmenting the CIFAR-10~\cite{krizhevsky2009learning} test set.
The new samples were generated by zooming (50\%) into the center of the images and saving intermediate stages (10 per sample).
While the data augmentation used might not replicate real-world scenarios, it demonstrates the feasibility of the decision mechanisms for image data.
BranchyNet's~\cite{teerapittayanonBranchyNetFastInference2016a} \ac{EE}-version of AlexNet was used for the inference to maintain comparability to the state-of-the-art research on \acp{EENN}.
The final classifier achieved an accuracy of 62\% on the augmented data.

\begin{figure}[ht]
\centering
\fontsize{10}{12}\selectfont
\includegraphics[width=1.0\linewidth]{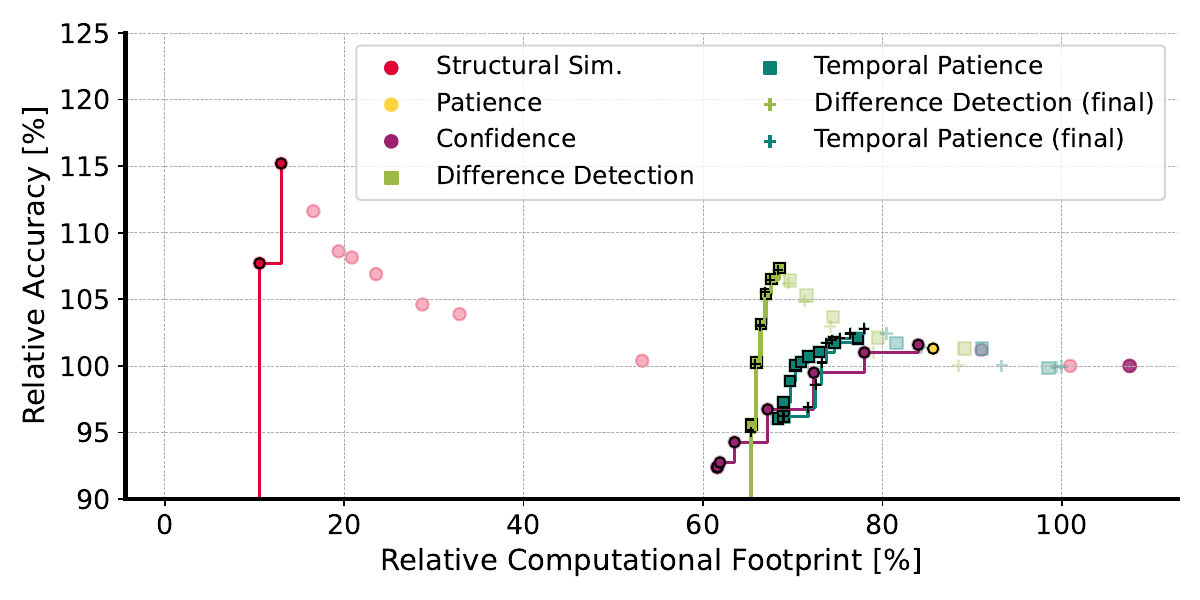}
\caption{Mean operations per inference vs. accuracy on the augmented CIFAR-10 test set for the different decision methods across different threshold configurations relative to the performance of the single exit version of the model.}
\label{fig:cifar10_scatter}
\end{figure}

Unlike input filtering, the Difference Detection and Temporal Patience approaches do not require a domain expert to create similarity metrics.
For image data, structural similarity~\cite{wang2004image} was used to quantify the change compared to a previous reference sample.
If the input is similar enough, no inference will be executed, and the previous output will be reused.
Although input filtering outperformed all \ac{EENN} decision mechanisms in terms of computational load (see Fig.~\ref{fig:cifar10_scatter}), it requires a larger memory footprint, as it must hold the reference sample in memory.
The temporal decisions were able to outperform the state-of-the-art decision mechanisms on this task, highlighting their usability for scenarios without the possibility of handcrafted input filtering solutions.

\noindent

\subsection{Speech Command Detection}


In the final experiment, we address a significant Edge AI application: wake-word detection.
This investigation focused on the Google Speech Commands (GSC) dataset~\cite{warden2018speech}.


Correlated test data was created by combining GSC test set samples into longer recordings, then split into input-sized segments with 0.9 second overlap.
Five-minute artificial recordings were used to reproduce typical wake word detection scenarios with lower-level white noise interrupted by 100 random noise events and 5 random commands.
Smaller overlaps resulted in reduced performance with the original model due to a lack of alignment between samples and commands.

We used an \ac{EENN} variant of the largest ``Hello Edge'' depthwise-separable convolutional architectures~\cite{zhangHelloEdgeKeyword2018a}, with three added \ac{EE}.
The \acp{EE} consist of a global pooling step, a dense layer, and the final activation function.
The first classifier is positioned immediately after the initial convolutional layer.
For this experiment, the Softmax activation layers were removed after training, allowing for a wider range of possible threshold configurations.
This change aimed to enhance the decision mechanism's sensitivity and reduce inference costs.



The decision to place the first \ac{EE} immediately after the initial convolutional layer was made to improve computational efficiency.
This was based on the assumption that the Difference Detection mechanism could benefit from less accurate but significantly cheaper \acp{EE}.



The classifiers achieved accuracies of 30.4\%, 88.2\%, 96.88\%, and 96.2\% on the augmented test recordings.
Notably, the first \ac{EE} in the shallow position exhibited a significant performance drop compared to the rest.
This \ac{EENN} is the only architecture in our evaluation that shows slight overthinking, as the last \ac{EE} demonstrated higher accuracy than the network's final classifier.


The decision mechanisms were benchmarked against confidence-based and patience-based solutions.
The patience-based approach was tested with decision thresholds of two and three, meaning that the last two or three classifier results needed to agree on the class label to trigger an early termination.

\begin{figure}
    \centering
    \includegraphics[width=1.0\linewidth]{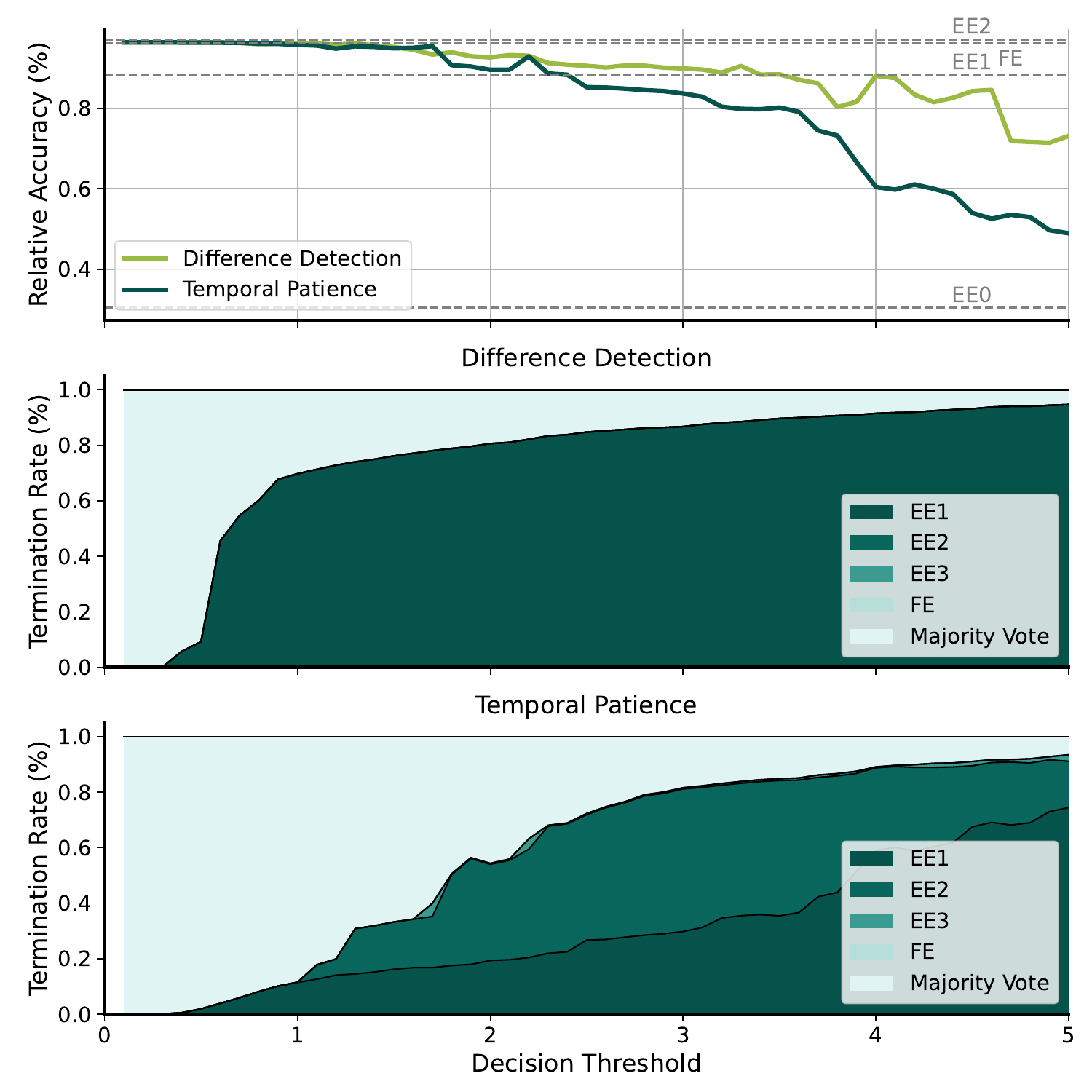}
    \caption{
    Comparison of accuracy between the Difference Detection and Temporal Patience mechanisms and sample termination rates at each classifier based on decision threshold for speech command detection.
    }
    \label{fig:gsc_share}
\end{figure}

The evaluation reinforced the observation from previous experiments, demonstrating enhanced efficiency outcomes at comparable accuracy levels when compared to state-of-the-art solutions.
However, unlike prior experiments, the Difference Detection solution showcased significantly smaller computational footprints while maintaining accuracy levels equivalent to those of the Temporal Patience approach.
This discrepancy arises from the initial \ac{EE}'s limited accuracy.
Integrating class labels as an auxiliary criterion for scene change detection limited the Temporal Patience approach's ability to terminate on the initial \ac{EE}.
The low accuracy of this \ac{EE} leads to fewer instances where its prediction is correct and aligns with the majority vote.
In contrast, the Difference Detection mechanism remains independent of class labels and can terminate its inference on up to 80\,\% of samples early while sustaining accuracy levels surpassing 90\,\%. Fig.~\ref{fig:gsc_share}  displays the accuracy and distribution of classifiers at different decision thresholds.
Furthermore, the initial \ac{EE}'s constrained accuracy diminishes the achievable efficiency gains of the confidence- and patience-based solutions as well.


\begin{figure}[ht]
\centering
\fontsize{10}{12}\selectfont
\includegraphics[width=1.0\linewidth]{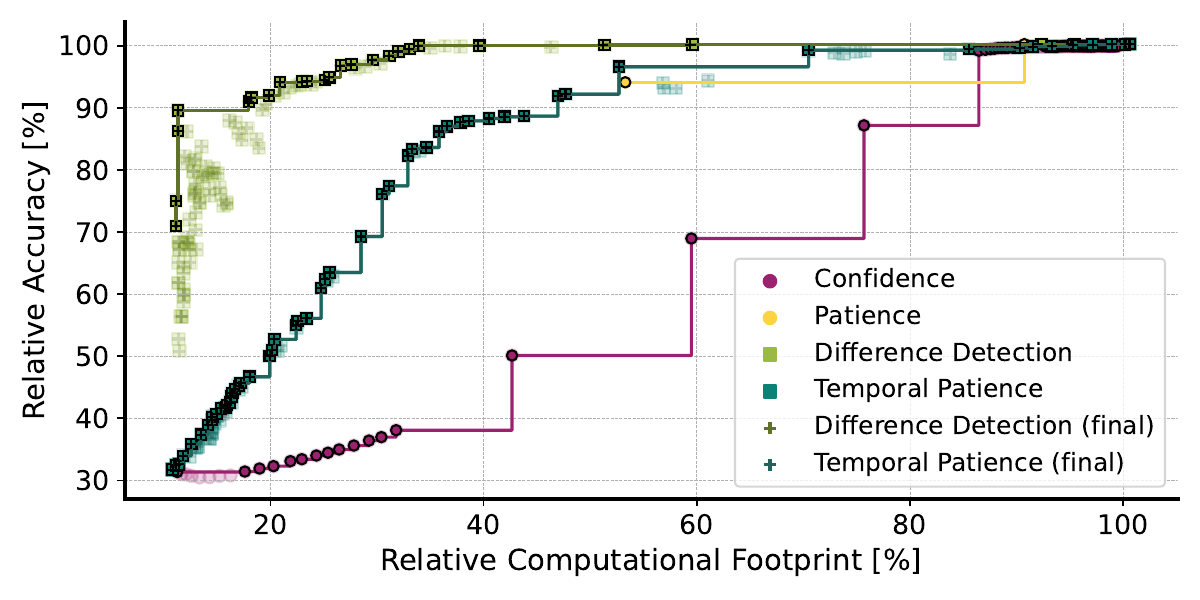}
\caption{
Mean operations per inference vs. accuracy relative to the single exit version of the model on the correlated GSC test data.
}
\label{fig:gsc_scatter}
\end{figure}

Another noteworthy insight emerges from the distribution of configurations depicted in the scatter plot (refer to Fig.~\ref{fig:gsc_scatter}): in contrast to state-of-the-art solutions, which exhibit a concave curve indicating that minor accuracy enhancements come at substantial computational costs, the curves associated with Difference Detection and Temporal Patience are convex.
This results in more gradual cost increments for accuracy improvements until a saturation point is reached.


This experiment underscores the potential efficiency gains achievable through both decision mechanisms in audio classification tasks.
Additionally, it highlights the prospect of further efficiency enhancements via optimized \ac{EENN} architectures tailored to these temporal decision mechanisms.

\noindent


\subsection{Evaluation of New Scene Labeling} \label{sec:new_scene}

The temporal decision mechanisms employ a majority vote to label the initial sample of each new scene.
This approach aims to mitigate the influence of overthinking. 
However, as only one of the experiments showed a (slight) overthinking behavior, we also evaluated the mechanisms with a different scene-labeling solution.
Instead of using the majority vote of all classifiers, only the output of the final classifier is considered.

This change results in only utilizing the first and last classifier for the Difference Detection.
Other branches can be removed, reducing the cost of labeling a new scene and the size of the model.

The Temporal Patience approach has no direct cost savings as it needs to execute all classifiers for a new scene to select the appropriate output for the subsequent samples.

The used \ac{NN} for the myocardial infarction detection only contains a single \ac{EE}.
This means that both labeling methods are identical for Difference Detection.
Temporal Patience does not show any improvement and only minimal changes in behavior.
The Difference Detection mechanism shows a marginal improvement when applied to the augmented CIFAR-10 dataset.
However, an increased computational footprint is observed for the Temporal Patience mechanism.
No substantial differences were detected between the two labeling methods for the speech commands use case.

This highlights that the labeling strategy for newly detected scenes holds negligible significance in the assessed scenarios.
Nevertheless, when implementing these mechanisms in novel applications and \ac{NN} architectures, the scene-labeling approach might require careful consideration depending on how prone to overthinking they are.


\section{Conclusion and Future Work}

In conclusion, this paper demonstrates the viability of harnessing temporal correlations within input data to guide the termination decisions of \acp{EENN} across diverse data modalities outside of highly correlated radar data.
The approach capitalizes on this correlation to reduce the computational load of the inference, achieving efficiency gains of up to 80\,\% while sustaining accuracy levels within 5\,\% of the original model.
These improvements will result in reduced mean latencies and extended battery life, enabling the deployment of more complex neural networks to \ac{IoT} devices.


The approach's key innovation is leveraging \ac{EE} results as simple embeddings to quantify input similarity, eliminating the need for domain-specific metrics and the storing of large reference inputs.
This enables the system to adapt to different data modalities and tasks.
While currently used for change detection, future work could explore other applications like monitoring solutions and improve the overall implementation by evaluating alternative distance metrics.
It is important to note that the method requires temporal correlation among input data.
However, given the prevalence of static and wearable sensors in \ac{IoT} applications, this property is present in the majority of applications within the embedded field.

Another important topic for future research is the optimal configuration of the used decision mechanism, its threshold hyperparameter and new scene handling.
This research, however, will require public datasets that contain subsets to represent the data to be expected in deployment scenarios, including correlation between subsequent samples and class distributions that are different from the equal distribution in training and test sets.
Such datasets could enable more effective evaluation of adaptive techniques, accelerate the development of new approaches, and lead to more useful solutions for real-world applications.


\begin{acks}
The project “RadarSkin” has received funding from the \ac{BMBF} under the call “Electronic Systems for Edge Computing” (grant number 16ME0543). The responsibility for the content of this publication lies with the author.
\end{acks}

\bibliographystyle{IEEEtran}
\bibliography{draft}

\end{document}